\documentclass[a4paper,conference]{IEEEtran}
\usepackage{cite}

\ifCLASSINFOpdf
  \usepackage[pdftex]{graphicx}
  \graphicspath{{./graphix/pdf/}{./graphix/jpeg/}}
  \DeclareGraphicsExtensions{.pdf,.jpeg,.png}
\else
\fi
\usepackage{amsmath}
\newcommand{\at}{\makeatletter @\makeatother}
\usepackage[inline]{enumitem}
\usepackage[colorinlistoftodos]{todonotes}

\begin{document}
%
\title{SRVIO: Super Robust Visual Inertial Odometry for Dynamic Environments and Challenging Loop-closure Conditions}

\author{\IEEEauthorblockN{Ali Samadzadeh}
\IEEEauthorblockA{ Computer Engineering Department\\
Amirkabir University of Technology\\
Tehran, Iran\\
Email: a\_samad\at aut.ac.ir}
\and

\IEEEauthorblockN{Ahmad Nickabadi}
\IEEEauthorblockA{ Computer Engineering Department\\
	Amirkabir University of Technology\\
	Tehran, Iran\\
	Email: nickabadi\at aut.ac.ir}





}


%


\maketitle

\begin{abstract}
There has been extensive research on visual localization and odometry for autonomous robots and virtual reality during the past decades. Traditionally, this problem has been solved with the help of expensive sensors, such as lidars. Nowadays, the focus of the leading research in this field is on robust localization using more economic sensors, such as cameras and IMUs. Consequently, geometric visual localization methods have become more accurate in time. However, these methods still suffer from significant loss and divergence in challenging environments, such as a room full of moving people. Scientists started using deep neural networks (DNNs) to mitigate this problem. The main idea behind using DNNs is to better understand challenging aspects of the data and overcome complex conditions such as the movement of a dynamic object in front of the camera that covers the full view of the camera, extreme lighting conditions, and high speed of the camera.  Prior end-to-end DNN methods have overcome some of these challenges. However, no general and robust framework is available to overcome all challenges together. In this paper, we have combined geometric and DNN-based methods to have the generality and speed of geometric SLAM frameworks and overcome most of these challenging conditions  with the help of DNNs and deliver the most robust framework so far. To do so, we have designed a framework based on Vins-Mono, and show that it is able to achieve state-of-the-art results on TUM-Dynamic, TUM-VI, ADVIO, and EuRoC datasets compared to geometric and end-to-end DNN based SLAMs. Our proposed framework could also achieve outstanding results on extreme simulated cases resembling the aforementioned challenges. 
\end{abstract}


%
\IEEEpeerreviewmaketitle

\section{Introduction}
\label{introduction}
Visual inertial odometry (VIO) is the process of determining the location and orientation of an entity with the help of visual (camera) and inertial (IMU) sensors. Today, visual odometry is being used in a wide variety of applications, including autonomous driving \cite{an2017semantic, sabry2019ground}, controlling and navigating aerial micro autonomous vehicles and nano-drones \cite{lin2018autonomous, delmerico2018benchmark, do2019high}, augmented reality and enabling navigation on mobile devices \cite{wu2015square}. In this paper, it is assumed that the target entity only has a single RGB camera and a single IMU sensor. 

The general pipeline of a modern VIO framework consists of three major parts\cite{qin2018vins}: preprocessing, optimization, loop closure and pose graph optimization. The first component (preprocessing) typically consists of visual feature tracking and IMU preintegration between frames. There are direct \cite{von2018direct, xiong2019ds}, keypoint based \cite{qin2018vins, campos2021orb} and deep-learning-based \cite{clark2017vinet, yang2020d3vo} visual feature tracking methods.

\begin{figure} 
\centerline{\includegraphics[width=\columnwidth]{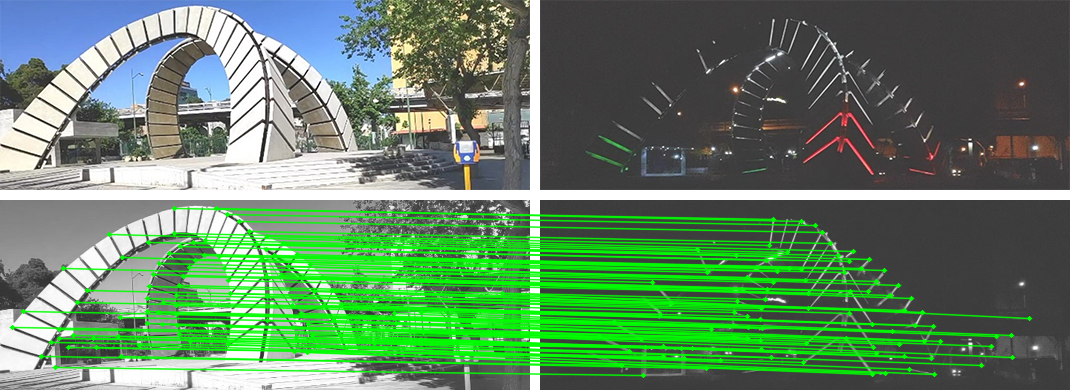}}
    \caption{An example of SRVIO framework’s loop-closure performed on  day and night images of the same scene. The loop is successfully discovered by SRVIO based on the matched keypoints. }
    \label{fig:front_page}
\vskip -0.2in
\end{figure}

The optimization component typically starts with an initialization stage in which, algorithms like five-point \cite{nister2004efficient} and triangulation are exploited to create an initial 3D point cloud and reconstruct the initial poses. The process is also known as visual Structure-from-Motion (SfM) \cite{schonberger2016structure}. After initial SfM, the extracted poses and 3D point cloud locations are fine-tuned using bundle adjustment and then the visual data are matched with inertial data by estimating the scale and initial gravity vector. 
Finally, the optimization process starts based on visual and inertial residuals. These residuals are prominent in state-of-art VIO frameworks \cite{xiong2019ds, qin2018vins, mur2017orb}. 

Loop closure and pose optimization is the last standard component of VIO frameworks. This component's goal is to globally optimize the poses using the information provided by loop closure and map reuse mechanisms. The loop closure mechanism is a tool for reducing the long-term error and improving robustness of the models and is claimed to be a crucial part of modern SLAM frameworks \cite{campos2021orb}. The three aforementioned components of a VIO system may be executed in parallel. There should only be inter-process communications for transferring data streams between these components.

Despite the success of the current state-of-the-art three-component VIO models, they fail dramatically in some challenging situations.  For example, the presence of dynamic objects in the field of view of camera  violates the scene rigidity assumption of the base VIO frameworks and degrades the visual features tracking quality. The moving objects make it difficult to build a stable and accurate 3D point cloud and poses. There have been many attempts, such as DynaSlam \cite{bescos2018dynaslam}, to extend monocular and stereo VIO to dynamic scenarios.

Also, VIO methods often struggle to keep track of visual features when the entity being tracked is moving quickly, resulting in blurry or completely new frames. Another major challenge is the existence of external objects that cover most of the camera's field of view and generate false visual cues of movement. In such cases, the visual odometry either provides wrong estimates or resorts to movements priors like assuming zero- or constant- speed movement.

Stable long-term localization is another challenge of VIO as there are issues with inertial sensors and visual tracking. Regarding the inertial sensors, integration over time diverges due to the accumulation of errors in measurements. On the other hand, the loop-closure, which plays a vital role in stabilizing VIO systems in long time usages, fails to match photos of the same scene taken at different times (e.g., day and night) or viewing angles (e.g., entrance and exit directions of an area). These situations are almost impossible for classical K-Nearest Neighbor (KNN) feature match and Perspective-n-Point to handle. While there are other deficiencies and limitations, we focus on the aforementioned issues in this paper.  

In this paper, we propose a novel VIO-framework to address the previously mentioned shortages of current state-of-the-art VIO methods. In this framework, a hybrid visual preprocessing step is designed to overcome the problem of the dynamic objects at real-time speed. Our proposed optimization component is capable of switching between different choices of useful data sources (including visual, inertial or both of them) to stay stable and provide accurate estimations under any circumstance. Our model's inertial preprocessing component solves the integration output divergence problem while it provides real-time output. Finally, we propose a novel transformer based loop-closure mechanism for robust localization in day-night and enter-exit scenarios. The proposed framework is evaluated against all of the state-of-the-art methods using well-known benchmarks and other challenging scenarios described in Section \ref{results}, and the results indicate the efficiency of the proposed solution.

The rest of this paper is structured as follows. The next section reviews the related works. Section \ref{framework} gives a full description of the proposed SRVIO model. Section \ref{results} reports the results of the experiments. Finally, Section \ref{conclusion}) concludes the paper.

\section{Related works}
\label{related}
This section will discuss the literature on VIO frameworks and the methods addressing the challenges reported in the previous section. First, in Section \ref{related:vio}, the most recent VIO frameworks utilizing a single camera and a single IMU are reviewed in detail. In Section \ref{related:dynamic}, methods that deal with moving objects and dynamic environments will be discussed. In Section \ref{related:IMU}, the problem of inertial data integration and odometry divergence is explained. Also, the frameworks that have tackled this problem are described. Lastly, the old problem of robust relocalization in challenging scenarios and the solutions provided for these situations are discussed in Section \ref{related:loopclosure}.

\subsection{Single-camera single-IMU VIO frameworks}
\label{related:vio}
There are many robust and successful frameworks that have been designed for single-camera single-IMU visual-inertial odometry, which can be divided into two categories: geometric and neural network-based methods. The former relies on traditional visual feature extraction methods, while the latter utilizes deep neural networks for feature extraction. One of the most famous geometric VIO frameworks is VINS-Mono\cite{qin2018vins}. VINS-Mono’s preprocessing steps are well-known algorithms; keypoints are extracted using the good-features-to-track algorithm \cite{shi1994good} and are tracked in video frames using Lucas-Kanade \cite{bouguet2001pyramidal} method. The optimization and initialization components of VINS-Mono are specially designed for this framework and play an essential role in the success of this framework. Despite the speed and accuracy of VINS-Mono in usual scenarios, it is not very stable in some challenging environments.

ORB-SLAM3 is another state-of-the-art geometric VI-SLAM framework that is more robust than VINS-Mono, addressing camera blur and visual feature loss and IMU-camera translation matrices error. However, the algorithm does not provide reliable outputs when dynamic objects appear in the scene despite its robustness. The loop-closure of ORB-SLAM is similar to the classic perspective-n-point and descriptor check method, which has its downsides when the loop-closure is required in day-night and enter-exit scenarios.

VI-DSO \cite{von2018direct} is another well-known geometric VI-SLAM algorithm that is based on sparse and direct visual feature tracking between frames and does not extract keypoints. This framework is also fast and accurate but has the same shortcomings as other geometric SLAM frameworks. Generally, geometric SLAM algorithms are fast, hand-engineered and robust to standard conditions, but they fail to provide a robust localization in extreme conditions stated in the previous section.

Recently, there has been an increasing interest in developing DNN-based SLAM frameworks. UnDeepVO \cite{li2018undeepvo} was the first visual odometry framework based on DNNs. The first attempts only proved the concept; they did not have the same accuracy and robustness as the geometric SLAMs. However, more recent frameworks such as VINet \cite{clark2017vinet} and D3VO \cite{yang2020d3vo} show promising results. VINet is a VIO model that formulates motion estimation as a sequence-to-sequence problem and uses an end-to-end network to learn it. The visual preprocessing is done by a FlowNetC NN in which the optical flow between frames is extracted, and the inertial preprocessing for integrating IMU data is accomplished via a LSTM network. VINet has achieved state-of-the-art performance but still suffers from robustness issues in the challenging environments mentioned in Section~\ref{introduction}. 

D3VO is a visual-only method that is more stable than VINet. The stability of this approach comes from a NN that estimates the depth of each frame and outputs the uncertainty coefficient of each pixel. This depth estimation NN uses the Bayesian filtering method introduced in \cite{liu2019neural} to solve the problem of image depth estimation in occluded and dynamic environments. The model shows outstanding performance and accuracy over various datasets. D3VO is not VIO-based and does not have a loop closure mechanism. Therefore, it diverges in long-term scenarios and could fail in other challenging situations like objects covering camera view or extreme motion blur.

\subsection{Dynamic robust SLAM frameworks}
\label{related:dynamic}
Today, almost all SLAM and VIO models are equipped with mechanisms to deal with non-rigid environments. One of the most straightforward methods is to use RANSAC in order to detect outliers\cite{mur2015orb, mur2017orb}.
Another reasonably simple method is to use a robust error function \cite{klein2007parallel}. Unfortunately, these approaches fail when the majority of keypoints belong to moving objects.

The more complex models tackle dynamic objects by identifying and dismissing dynamic keypoints or dynamic objects entirely.
To identify moving objects, the common approach is to use semantic segmentation to remove the areas belonging to dynamic objectt classes.
Mask-SLAM \cite{kaneko2018mask} uses a mask produced by a semantic segmentation algorithm to exclude undesirable feature points.
DS-SLAM \cite{yu2018ds} is equipped with a real-time semantic segmentation network running in an independent thread which coupled with moving consistency checking methods, enables it to filter out the dynamic portion of the scene.
Similarly, PSPNET-SLAM \cite{han2020dynamic} uses a semantic segmentation network (PSPNET \cite{zhao2017pyramid}) combined with ORB-SLAM2 to achieve the same goal.
RGB-D SLAM \cite{li2017rgb} filters out dynamic keypoints instead of objects by only using depth edge points for visual odometry. These depth edges have static weights that indicate each point's probability of belonging to a static object.

Dyna-SLAM \cite{bescos2018dynaslam} applies pixel-wise segmentation using a CNN to detect dynamic objects for the monocular and stereo cases. In the RGBD case, multi-view geometry and deep learning are used to remove dynamic objects and in-paint the background occluded by these objects.
Finally, Dynamic-SLAM \cite{xiao2019dynamic} consists of an SSD (Single Shot Detector) object detector for detecting dynamic objects, and a missed detection compensation algorithm to improve the recall rate of the SSD component, and a feature-based visual SLAM that correctly handles the feature point of the dynamic objects. 

Most of the methods mentioned above achieve real-time dynamic object filtering. Nevertheless, when the dynamic object covers the whole field of view in extreme cases, the frameworks' outputs diverge or show motion prior. This paper shows that using another information source like the inertial sensor's data can help to correct motion estimation errors in these extreme situations.

\subsection{Inertial Odometry}
\label{related:IMU}
Integrating inertial data (linear acceleration and rotation speed) provided by the IMU sensor is an excellent odometry method for short periods. However, in long term, the output diverges dramatically, as discussed and demonstrated in \cite{woodman2007introduction}. Some methods, such as \cite{savage1998strapdown}, try to limit IMU integration error over time, but they are not very useful because they are designed for particular applications such as pedestrian movements and are not very accurate as well. With the advances in deep learning methods in processing time-series data, new methods emerged to solve this problem known as inertial odometry. 

One of the first successful algorithms which used DNNs to cure the curse of drift in inertial odometry was IONet \cite{chen2018ionet}. This method, with proper initialization, performed close to industrial VIO frameworks such as Google Tango. IONet uses an LSTM-based architecture and polar coordinates vectors to estimate the pose at each step. It outperforms the previous classical attempts by a large margin. 

The other idea was the use of denoising CNNs. For example,  \cite{brossard2020denoising} tries to remove the noise that causes the integration to drift. Such methods require calibration data from a stationary entity with the IMU sensor. Almost all new emerging VIO frameworks have a performance gap against modern VIO and VO frameworks, but fusing them with geometric SLAM’s preprocessing units can show promising results over challenging scenarios. In this paper, we have designed a denoising CNN to do the aforementioned task and have combined it with a geometric SLAM (details provided in section \ref{framework}).

\begin{figure*} 
\includegraphics[width=\textwidth]{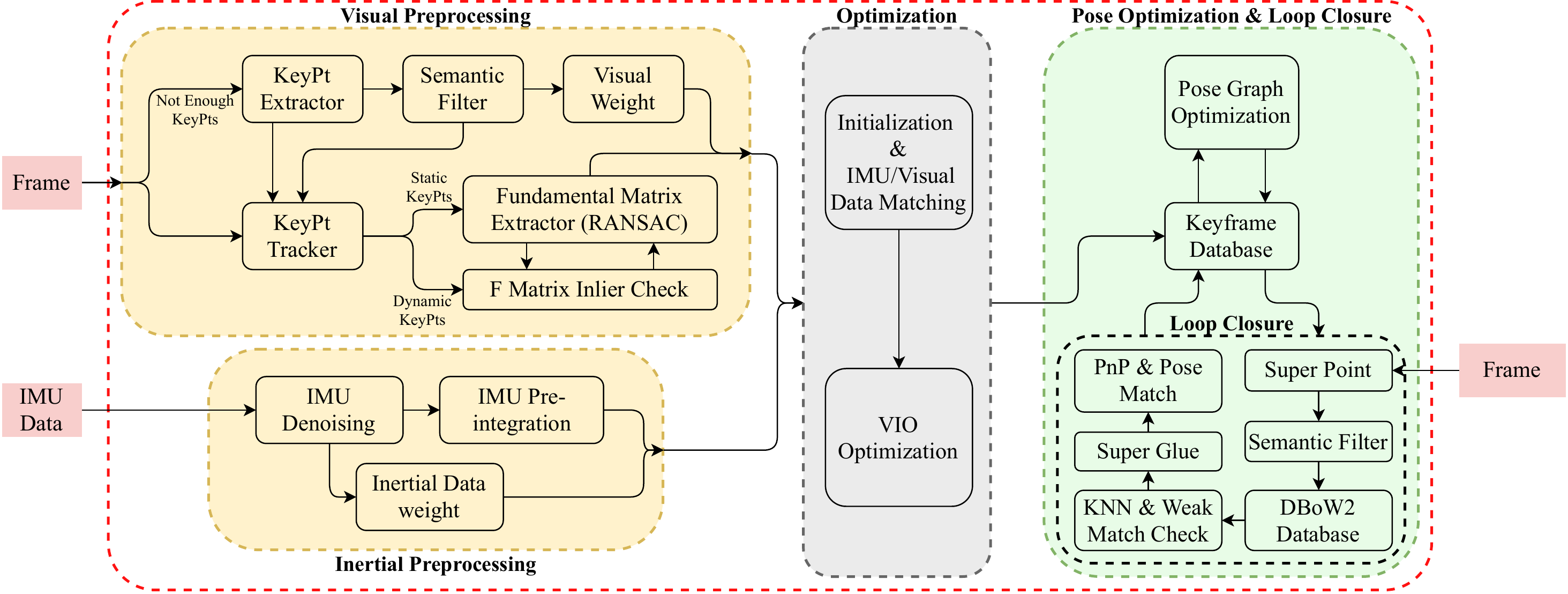}
   \caption{SRVIO framework architecture} 
   \label{fig:architecture}
\end{figure*}

\subsection{Robust Relocalization}
\label{related:loopclosure}
Robust relocalization over extreme scenarios such as day-night, rainy or snowy weather, seasonal changes, extreme viewpoint changes, and similar scenarios is a big challenge in VIO. So far, many methods have tried to overcome some of these challenges. The most interesting recent methods are X-view \cite{gawel2018x} and Superglue \cite{sarlin2020superglue}. The X-view method uses semantic segmentation over frames to build a graph of objects over frames; then, relocalization is done via sub-graph matching. The method is interesting because it can solve almost all relocalization challenges, especially the extreme viewpoint changes. The method can even relocalize the same place seen on the ground from aerial images. However, this method is not accurate and gives a rough estimation of the location and not a transformation matrix. 

The Superglue provides exact transformation between two frames of a video. This method is based on Superpoint \cite{detone2018superpoint} visual keypoint extractor and descriptor proposed by the same team. It uses graph transformers to find appropriate matches between the corresponding keypoints in frames and shows robustness over day-night conditions and rainy weather, but it does not show robustness over seasonal changes. Despite the success of Superglue in accurate estimation of transformation between images with robustness over extreme changes, it has never been used in a classical loop closure mechanism inside geometric SLAM. In this paper, we use the Superglue and Superpoint inside a novel loop closure mechanism to improve the relocalization robustness of our model.

\section{SRVIO framework}
\label{framework}
The proposed SRVIO framework is explained in detail in this section. As demonstrated in Fig. \ref{fig:architecture}, the SRVIO consists of three main components: preprocessing, optimization and loop closure. The input to this framework is the image stream of a single RGB camera and IMU sensor data. The final output is the pose of the camera in each key frame and the corresponding 3D point cloud. In addition to the poses and locations, each frame's keypoints are reported in the form of pixel locations and their descriptors. In the following subsections, each component of the model is discussed in detail. 

\subsection{Preprocessing}
\label{sec:framework:preprocessing}
The preprocessing module of SRVIO is made up of two visual preprocessing and inertial preprocessing blocks, as seen in Fig.~\ref{fig:architecture}.\\

\subsubsection{Visual preprocessing}

The visual preprocessing block of our SLAM system takes in a sequence of frames ($I_i$) and outputs corresponding keypoints matrices ${\mathcal{P}}^{\mathcal{C}_i}$ between each pair of consecutive frames, as well as a weight for each frame which represents the quality of the keypoints in that frame.
To do this, it first employs a keypoint extractor, good-features-to-track \cite{shi1994good}, to track previous frame's keypoints and extract new keypoints from the regions of frame $i$ in which the old keypoints of frame $i-1$ do not exist. The result is the set of initial keypoints of current frame as ${\mathcal{P}}^{{\mathcal{C}_i}}_{initial} = GoodFeaturesToTrack(I_i, {\mathcal{P}}^{\mathcal{C}_{i-1}})$. 

If keypoints extracted from the regions occupied by non-stationary objects are used, they will provide misleading clues for motion estimation. Hence, the initial keypoint set is then matched against the output of a semantic segmentation neural network (e.g., HRNet \cite{yuan2020object}) to decide whether they are possibly dynamic (like keypoints on humans, cars, and trains) or static (like keypoints on buildings or streets). This sub-step is known as the semantic filter in our model. The semantic segmentation network segments the input image into non-overlapping regions labeled as different items of a predefined set of objects. The output of the semantic segmentation algorithm is then converted to a binary mask indicating static and dynamic regions of the input frame. The dynamic region is then dilated to contain the borderline pixels, preventing the model from defining keypoints on these lines. 

The semantic segmentation network can be implemented inside a SLAM framework by various approaches, as in other methods like \cite{schorghuber2019slamantic} and \cite{bescos2018dynaslam}. The output of this network is used for filtering out the undesired (dynamic) keypoints. However, it is also possible to use the semantic segmentation output before the keypoint extraction step to restrict the search for keypoints to the stationary regions of the input frame. However, this may cause the model to be unstable as there might be many potentially dynamic objects (e.g., humans) that are already static in the scene and rejecting all of the keypoints on these objects remains very few keypoints for proper tracking and optimization.

\begin{figure} 
\centerline{\includegraphics[width=\columnwidth]{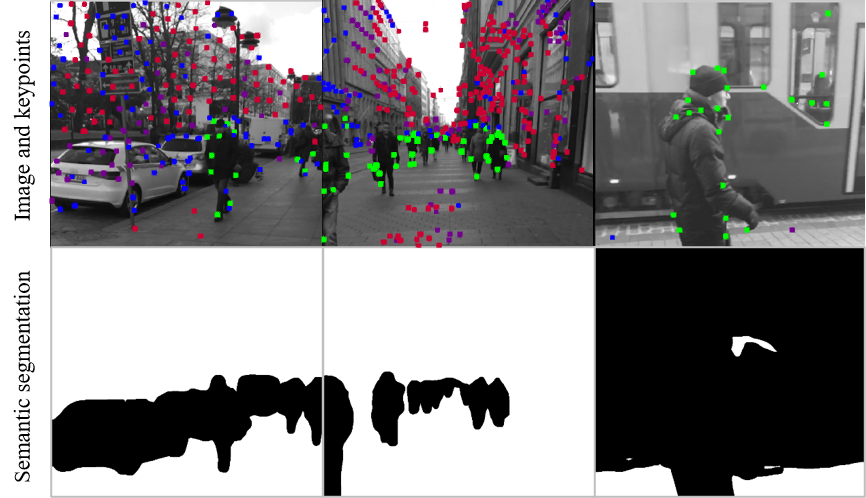}}
    \caption{Examples of the effect of semantic segmentation on keypoint selection. Top: three frames from seq22 of the Advio dataset on which the keypoints detected on the dynamic objects are shown in green. Bottom: dynamic semantic mask extracted for each frame. The figures show that the keypoints on steady cars are not removed in the proposed method.}
    \label{fig:ss_example}
\end{figure}

The keypoint selection mechanism of SRVIO is fast and accurate. It is fast since the semantic segmentation neural network is employed only when new keypoints are required due to objects getting out of the scene or being covered. It is also accurate since it allows the use of the keypoints wrongly labeled as dynamic, e.g., points on steady cars. To do this, SRVIO employs a three-step algorithm. First, RANSAC \cite{fischler1981random} is applied only to the keypoints labeled as static by the semantic segmentation network to find initial fundamental matrix transformation ($F^{i,i-1}_{inital}$) between frame $i$ and $i-1$. Next, a test is performed to check the consistency of all keypoints, including those labeled as dynamic objects by the segmentation network, with this initial fundamental matrix and only the truly moving keypoints are removed. Finally, the RANSAC is again applied to the new set of keypoints. These three steps can be formulated as follows:

\begin{align}
\left[ {\mathcal{P}}^{\mathcal{C}_{i, S1}}, F^{i,i-1}_{inital} \right] &= RANSAC \Big( {\mathcal{P}}^{\mathcal{C}_i}_{initial}, {\mathcal{P}}^{\mathcal{C}_{i-1}} \Big) \\
{\mathcal{P}}^{\mathcal{C}_{i, S2}} = \Bigg\{ {\mathcal{P}}^{\mathcal{C}_{i}}_{l, initial} & : \ \left\lvert {\mathcal{P}}^{\mathcal{C}_{i}}_{l, initial} . F^{i,i-1}_{inital} . {\mathcal{P}}^{\mathcal{C}_{i-1}}_l \right\lvert < \epsilon \Bigg\} \label{eq:f_limit}\\
\left[ {\mathcal{P}}^{\mathcal{C}_{i}}, F^{i, i-1}_{accurate}\right] &= RANSAC \Big( {\mathcal{P}}^{\mathcal{C}_{i, S2}}, {\mathcal{P}}^{\mathcal{C}_{i-1}} \Big) \label{eq:acc_ransac}
\end{align}
where ${\mathcal{P}}^{\mathcal{C}_i}_{initial}$ is the initial keypoint extracted from good-features-to-track algorithm. $F^{i,i-1}_{inital}$ and ${\mathcal{P}}^{\mathcal{C}_{i, S1}}$  are the fundamental matrix and the keypoint matrix corresponding to the initial static keypoints.
The Equation \ref{eq:f_limit} checks the fundamental matrix test with the corresponding keypoints of $i^{th}$ frame and static keypoints of ${(i-1)}^{th}$ frame. The output (${\mathcal{P}}^{\mathcal{C}_{i, S2}}$) is the new set of keypoints including the keypoints on static objects initially mislabeled as dynamic. Finally, in order to accurately obtain non-dynamic keypoints and remove the outliers and dynamic keypoints, the RANSAC algorithm is performed again in Equation \ref{eq:acc_ransac}. The ${\mathcal{P}}^{\mathcal{C}_{i}}$ is the final matrix of keypoints.
Fig. \ref{fig:ss_example} shows three examples of keypoint extraction and filtering in SRVIO along with the corresponding static/dynamic masks. The green, blue and red dots on the frames are filtered, tracked and not-tracked keypoints, respectively. The white (black) regions of the mask show static (dynamic) objects. As shown in this figure, the keypoints on the steady cars in the left picture are not filtered even though the semantic mask labels them as dynamic, while the keypoints on the train in the right picture are not used as expected.

After eliminating the moving keypoints, a weight parameter ($\Psi_{c_{i}}$) is computed for each frame that determines the quality of the keypoints of that frame. This frame weight is used in the optimization component and is defined as follows:
\begin{align}
	\Psi_{c_{i}} &=  \dfrac{\left| {\mathcal{P}}^{\mathcal{C}_{i}} \right| }{F_{max}}
\end{align}
where $|\,.\,|$ denotes the size of the set of all remained keypoints inside $i^{th}$ frame. The $F_{max}$ is a constant indicating the maximum number of keypoints that can be present in a frame (experimentally set to 250 in this paper). Accordingly, the frame weight is defined as the number of selected static points in proportion to the maximum points extracted by the keypoint extractor algorithm for a frame. This weight helps to decrease attention to the visual error where the visual clues are unreliable because the dynamic objects are covering the camera's view field or the keypoint extractor is unable to extract a proper number of keypoints due to motion blur or similar challenges inside frames. \\

\subsubsection{Inertial preprocessing}

\begin{figure} 
\centerline{\includegraphics[width=\columnwidth]{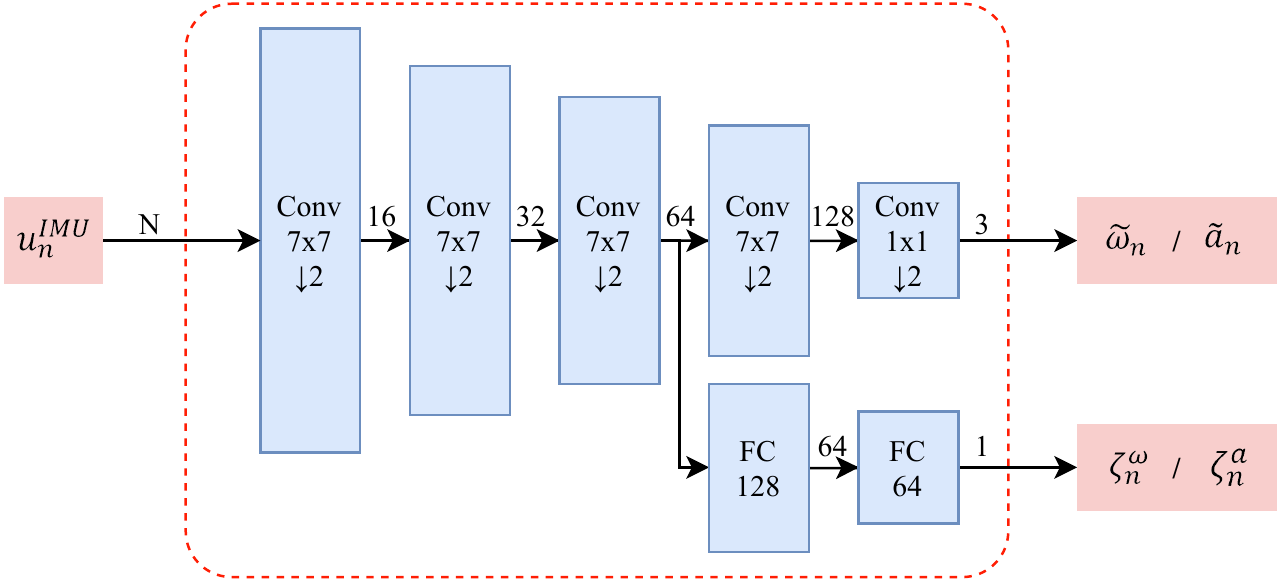}}
\caption{IMU denoising CNN architecture.}
    \label{fig:denoise_cnn}
\end{figure}

In the inertial preprocessing component, a denoising convolutional neural network (DCNN) is used to help eliminate unwanted IMU noise and mitigate the divergence issue. As shown in Fig. \ref{fig:denoise_cnn}, the proposed DCNN consists of two clone networks: one for gyro and the other for the accelerometer data denoising. The gyro DCNN gets $N$ sequential IMU data from time steps $i-N$ to $i$ and outputs gyro correction ($\tilde{\omega}_i$) for the $i^{th}$ IMU measurement. The accelerometer DCNN is the same as the gyro DCNN except that it is trained to estimate the accelerometer correction ($\tilde{a}_i$) for the $i^{th}$ IMU data. The correction fixes the IMU bias and random walk noise via separating the noise part of the noisy IMU data. The correction is performed as follows:
\begin{align}
\label{eqn:inertial_correction}
\hat{C}_{(.)} &= \hat{S}_{(.)}\hat{M}_{(.)} \nonumber \\
\hat{\omega}_{i}&=\hat{\mathbf{C}}_{\omega} \omega_{i}^{IMU}+\tilde{\omega}_{i} \\ 
\hat{a}_{i}&=\hat{C}_{a} a_{i}^{IMU}+\tilde{a}_{i} \nonumber \\
\zeta_i &= \zeta^{\omega}_i + \zeta^{a}_i \nonumber
\end{align}
where $\hat{M}_{(.)} $ are axis misalignment matrices and $\hat{S}_{(.)}$ are scale factors; therefore, $\hat{C}_{(.)}$ is the correction matrix of the raw input data. These parameters are unique for each IMU and are learned for each dataset. The $(.)^{IMU}_i$, $\tilde{(.)}_i$ and $\hat{(.)}_i$ correspond to IMU raw measurements, denoising correction values of the corresponding DCNN, and the final corrected values, respectively. $\zeta^{a}_i$ and $\zeta^{\omega}_i$ are the accelerometer and gyro data quality scores for $i^{th}$ IMU data. This score improves the framework's robustness over faulty IMU sensors and similar challenges.

The denoising networks of our model (Fig. \ref{fig:denoise_cnn}) are trained in the following manner. First, the convolution layers are trained using GELU activation function and cosine schedulers using the labeled IMU calibration data. Then, the convolutional layers are frozen and the linear layers are trained with augmented IMU data contaminated with different levels of random noise while the labels are the noise percentage. Consequently, the network is trained to predict the IMU data quality or the noise level in this stage. Moreover, since ground truth is not available with high frequency, the corrected IMU (gyro and accelerometer) data is integrated through time to reach the timestep interval of the ground truth data. The training loss is defined as follows:
\begin{align}
\delta \hat{R}_{i, i+j} &= \prod_{k=i}^{i+j-1} \exp \left(\hat{\omega}_{k}\right)\\
\delta \hat{P}_{i, i+j} &= \int \int_{t \in [i, i+j]} \exp \left(\hat{\omega}_{t}\right) \ \hat{a}_t \ dt^2 \\
\mathcal{L}^{\omega}_j &= \sum_i \rho\left(\log \left(\delta R_{i, i+j} \delta \hat{R}_{i, i+j}^T \right)\right) \\
\mathcal{L}^{a}_j &= \sum_i \rho\left(\delta P_{i, i+j} - \delta \hat{P}_{i, i+j} \right)\\
\mathcal{L}^{\omega} &= \mathcal{L}^{\omega}_{16} + \mathcal{L}^{\omega}_{32}\\
\mathcal{L}^{a} &= \mathcal{L}^{a}_{16} + \mathcal{L}^{a}_{32}
\end{align}
where $\mathcal{L}^{a}$ and $\mathcal{L}^{\omega}$ are the main loss functions for training accelerometer and gyro denoising networks, respectively. The $\hat{(.)}$ symbols represent corrected IMU measurements and $\delta P_{i, i+j}$ and $\delta R_{i, i+j}$ represent ground truth values of position and rotation in the time period of $i$ to $i+j$. $(.)^T$ is used to show the transpose of a matrix. Since ground truth of rotations is not available with high frequency for training accelerometer denoising network and calculation of $\delta \hat{P}_{i, i+j}$, this network is trained after the gyro denoising network training in order to have an accurate rotation matrix per data ($\exp \left(\hat{\omega}_{t}\right)$). The $\rho(.)$ is the Huber-norm defined in Equation \ref{eq:hubernorm}. The final loss of each network is defined as the sum of the losses for 16 and 32 time steps.

\begin{equation}
	\rho(s) = 
	\begin{cases}
	1 & s \geq 1, \\
	2\sqrt{s} - 1 & s < 1.
	\end{cases}
	\label{eq:hubernorm}
\end{equation}

Hereafter in SRVIO, corrected values of accelerometer and gyro are regarded as the IMU measurements on which the pre-integrations and integrations are performed with the assumption that bias and noise standard deviation are near zero. Using the IMU data validity score ($\zeta_i$), defined in Equation \ref{eqn:inertial_correction}, the IMU attention weight is defined as follows:
\begin{align}
\Psi_{b_i} = \frac{\sum_{k \in M_{l,j}} \zeta_k}{m}
\end{align}
where $M_{l,j}$ is the set of IMU data inputs between $l^{th}$ frame and $j^{th}$ frame, and $m$ is the size of this set. This weight parameter may be set to 1 if the user is confident about the IMU data quality.

\subsection{Optimization}
The location and pose estimation of SRVIO starts with an estimator initialization step and continues with an optimization-based VIO procedure combined with loop-closure detection. The initialization component of our model is similar to those of the previous frameworks like VINS-Mono whose details are given in \cite{qin2018vins}. Thus, in this section, we only focus on the details of our proposed optimization module. In the following, we first define the residual terms for visual and IMU measurements and then introduce the final error function. 

The formulas of this section use the following notations:
$(.)^w$, $(.)^b$, $(.)^c$ represent the world frame, the body frame (i.e. the IMU frame) and the camera frame, respectively.
$R$ and $q$ are used as matrix and quaternion of rotation, respectively.
The subscript for $R$ or $q$ indicates the original frame while the superscript denotes the destination frame. $p$ is used to show translation.
$b_k$ and $\mathcal{C}_k$ are used to show body frame and camera frame while taking the $k^{th}$ image.
$\bigotimes$ is used to denote multiplication operation between two quaternions.
$g^w = [0,0, g]^T$ is the gravity vector. 
Finally $\hat{(.)}$ denotes a noisy measurement.

\subsubsection{IMU measurement residuals}
Let us denote accelerator and gyroscope denoised values by $\hat{a}$ and $\hat{\omega}$. The IMU pre-integration terms are then defined as
\begin{align}
\alpha^{b_k}_{b_{k+1}} = \int &\int_{t \in [t_k, t_{k+1}]} R^{b_k}_t(\hat{a}_t)dt^2 \nonumber \\
\beta^{b_k}_{b_{k+1}} = &\int_{t \in [t_k, t_{k+1}]} R^{b_k}_t(\hat{a}_t)dt
\\
\gamma^{b_k}_{b_{k+1}} = &\int_{t \in [t_k, t_{k+1}]} \dfrac{1}{2}\Omega (\hat{\omega}) \gamma_t^{b_k} dt
\nonumber
\end{align}
where $\Omega(.)$ is:
\begin{align}
	\Omega(\omega) &= 
	\begin{bmatrix}
		- \big \lfloor \omega \big \rfloor_{\times} 	&&  \omega  \\
		-\omega^{T}										&&  0		\\
	\end{bmatrix}, 
	\nonumber \\
	- \big \lfloor \omega \big \rfloor_{\times} &= 
	\begin{bmatrix}
		0 		  && -\omega_z && \omega_y  \\
		\omega_z  && 0 		   && -\omega_x \\
		-\omega_y && \omega_x  && 0		
	\end{bmatrix}.
\end{align}

The residual for two IMU measurements in consecutive frames at body (IMU) coordinates $b_k$ and $b_{k+1}$ in the sliding window can be written as:
\begin{align}
	&r_{\mathcal{B}}(\hat{z}_{b_{k+1}}^{b_k}, \chi) = 
	\begin{bmatrix}
		\delta \alpha^{b_k}_{b_{k+1}} \\
		\delta \beta^{b_k}_{b_{k+1}} \\
		\delta \theta^{b_k}_{b_{k+1}} 
	\end{bmatrix}
	\nonumber \\
	&=
	\begin{bmatrix}
		R_{w}^{b_k} \big( p_{b_{k+1}}^w - p_{b_{k}}^w + \dfrac{1}{2} g^w\Delta t^2_k - v_{b_k}^w \Delta t_k\big)
		- \hat{\alpha}^{b_k}_{b_{k+1}} \\
		R_{w}^{b_k} \big( v_{b_{k + 1}}^w + g^w\Delta t_k - v_{b_k}^w \big) - \hat{\beta}^{b_k}_{b_{k+1}} \\
		2 \big[ {q^{w}_{b_k}}^{-1} \bigotimes q^w_{b_{k+1}} \bigotimes (\hat{\gamma}^{b_k}_{b_{k+1}})^{-1} \big]_{xyz} 	
	\end{bmatrix},
\end{align}
where $\Delta t$ is time interval, $v$ denotes velocity,  $[\,.\,]_{xyz}$ is an operator extracting the vector part of quaternion $q$ for error state representation and $\delta \theta^{b_k}_{b_{k+1}}$ is the three dimensional error state representation of the quaternion. When the DCNN is enabled, the IMU biases are not optimized and bias terms will be considered constant and zero.

\subsubsection{visual measurement residual}
The camera model defines camera measurement residuals on a generalized unit sphere in contrast to the traditional pinhole camera model. 
Considering the $l^{th}$ feature which appears at the $i^{th}$ image, the residual for observing that feature at the $j^{th}$ image is written as:

\begin{align}
&r_{c}(\hat{z}_{l}^{c_j}, \chi) = 
 [b_1 b_2]^T.(\mathcal{P}^{c_j}_l - \dfrac{\mathcal{P}^{c_i}_l}{\|\mathcal{P}^{c_i}_l\|}) &
\nonumber \\
&\mathcal{P}^{c_j}_l =
\pi_{c}^{-1}
(\begin{bmatrix}
u_l^{c_j} \\
v_l^{c_j} 
\end{bmatrix}) &
\\
&\mathcal{P}^{c_i}_l = 
R^{c}_{b}(
R^{b_j}_{w}(
R^{w}_{b_i}(
R^{b}_{c}
\dfrac{1}{\lambda_l}\pi_c^{-1}(
\begin{bmatrix}
u_l^{c_i} \\
v_l^{c_i}
\end{bmatrix})
+ p^{b}_{c}
) 
\nonumber
\\
& + p^{w}_{b_i} - p^{w}_{b_j} )- p^{b}_{c}), &
\nonumber
\end{align}
where
$\begin{bmatrix}
u_l^{c_i} \\
v_l^{c_i}
\end{bmatrix}$
denotes the first observation of $l^{th}$ feature in the $i^{th}$ image while 
$\begin{bmatrix}
u_l^{c_j} \\
v_l^{c_j}
\end{bmatrix}$
is the observation of that feature in the $j^{th}$ image.
$\pi_c^{-1}$ is the back projection function which given a pixel location yields the corresponding unit vector using camera intrinsic parameters. The residual vector is projected onto the tangent plane since vision residual has two degrees of freedom. $b_1$, $b_2$ are bases spanning the tangent plane $\mathcal{P}^{c_j}_l$.

\subsubsection{full error function}
The full state of window denoted by $\chi$ can be written as:
\begin{align}
\chi &= [ \mathbf{x}_0, \mathbf{x}_1, \dots \mathbf{x}_n, \mathbf{x}_c^b, \lambda_1,\ldots\lambda_m ],
\label{eq:m:2} \\
\mathbf{x_k} &= [p_{b_k}^w, v_{b_k}^w, q_{b_k}^w], \in [0, n],\\
\mathbf{x_c^b} &= [p_{c}^b, q_{c}^b],
\end{align}
where $\mathbf{x_k}$ is the IMU state when the $k^{th}$ image is captured. $p_{b_k}^w, v_{b_k}^w \text{ and } q_{b_k}^w$ denote position, velocity and the orientation of the body in the world frame at the time $k$, respectively. $n$ indicates the number of key frames in the sliding window while $m$ is the number of features in that window. Finally $\lambda_l$ is the inverse depth of $l^{th}$ feature in the sliding window from its first observation.

The base error function used in VINS-Mono \cite{qin2018vins} can then be written as:
\begin{align}
\mathcal{R} = \| r_p - H_p \chi  \|^2 &+ 
\sum_{k \in \mathcal{B}}
\| r_{\mathcal{B}}(\hat{z}_{b_{k+1}}^{b_k}, \chi) \|_{P_{b_{k+1}}^{b_k}}^2 \nonumber \\
&+ \sum_{l, j \in \mathcal{C}} \rho(\| r_{\mathcal{C}}(\hat{z}_{l}^{\mathcal{C}_j}, \chi) \|_{P_l^{\mathcal{C}_j}}^2),
\label{eq:m:1}
\end{align}
where $r_{\mathcal{B}}(\hat{z}_{b_{k+1}}^{b_k}, \chi)$ 
and $r_{\mathcal{C}}(\hat{z}_{l}^{\mathcal{C}_j}, \chi)$
are residuals for visual and IMU measurements, respectively, and $\mathcal{C}$ is the set of features which have been observed at least twice in the current sliding window. $r_p$ and $H_p$ are prior information from marginalization and $\rho(.)$ is the Huber norm (defined in Eqn. \ref{eq:hubernorm}).

The Huber function $\rho$ penalizes residuals contributed by outlier features, but it assigns equal importance to both camera and IMU residuals. However, camera measurements usually yield a more satisfying result when sufficient keypoints are present in the scene. IMUs are usually vulnerable to shifts over long periods, but they exhibit precise performance during short periods. The goal of the optimization component of SRVIO is to utilize the potential of each of the measurements to compensate for the deficiencies of the other measurement source in specific scenarios. To achieve this goal a novel weighted error-function is proposed as follows:

\begin{align}
\mathcal{R}_{2} = \| r_p - H_p \chi  \|^2 &+ 
\sum_{k \in \mathcal{B}}
\Psi_{b_k} \ \| r_{\mathcal{B}}(\hat{z}_{b_{k+1}}^{b_k}, \chi) \|_{P_{b_{k+1}}^{b_k}}^2 \nonumber \\
&+ \sum_{l, j \in \mathcal{C}} \Psi_{c_k} \  \rho(\| r_{\mathcal{C}}(\hat{z}_{l}^{\mathcal{C}_j}, \chi) \|_{P_l^{\mathcal{C}_j}}^2)
\label{eq:m:3}
\end{align}

This error function assigns a weight to each of the visual and inertial terms according to the number of keypoints present in the current window (Eqn. \ref{eq:m:3}). The camera residual weight increases in proportion to the number of visual keypoints in the frame window and causes the model to rely on the visual clues in windows with large number of keypoints. On the other hand, the error function mainly depends on IMU residual in frame windows with few keypoints.

\subsection{Loop closure}
The loop closure component is essential for the long-term accuracy of trajectory estimation as it corrects the large drifts that may occur over time. The algorithm needs to memorize the previous scenes it has visited to detect the loops. However, raw frames are not saved; instead, a bag-of-words representations of key frames is stored in a visual database to decrease memory usage. To do this, a set of keypoints are detected in each keyframe. Each keypoint is then described as a visual word using the Superpoints keypoint descriptor~\cite{detone2018superpoint}. Once a loop -i.e. the correspondence between the input frame and a frame in the keyframe database- is detected, the estimated path is optimized to match the detected loop.

As shown in Fig. \ref{fig:architecture}, the loop closure component consists of two blocks: loop closure detection block and pose graph optimization block which closes the loops and enables map reuse.
The loop closure component performs a yaw and translation optimization defined as below:

\begin{align}
\min_{p, \psi} & \Bigg\{ \sum_{(i,j) \in E}  \| \mathbf{r}_{i,j} \|^2  \Bigg\} \\
\bar{P}_{i,j} &= \bar{R}^{w^{-1}}_i \left( \bar{P}^w_j - \bar{P}^w_i \right)
\nonumber\\
\bar{\psi}_{i,j} &= \bar{\psi}_j - \bar{\psi}_i 
\\
\mathbf{r}_{i,j}(P^w_i, \psi_i, P^w_j, \psi_j) &= \begin{bmatrix}
R^{w^{-1}}_i \left( P^w_j - P^w_i \right) - \bar{P}_{i,j}\\
\psi_j - \psi_i - \bar{\psi}_{i,j}
\end{bmatrix}
\nonumber
\label{eq:loopresidual}
\end{align}
where $E$ is the set of all pose graph edges and $(\,\bar{.}\,)$ indicates a constant value. Additionally, $\bar{\psi}_{i,j}$ is the constant previous yaw difference between two graph nodes.

The VINS-Mono algorithm divides the equation \ref{eq:loopresidual} into loop closed edges and sequential edges to reduce the risk of wrong loop closure. However, the Superglue method is very robust, and most of the correspondences result in correct relocalization, and there is no need for such division.

The second block in this component is the loop closure detection block. The loop closure detection block is crucial for maintaining long-term trajectory estimation accuracy. This block should be robust to challenges such as light source, day/night, and viewpoint changes. Previous methods (e.g., VINS-Mono\cite{qin2018vins}) use some simple keypoint extractors and KNN mapping methods for this purpose which are useless in facing the aforementioned challenges. 
SRVIO, on the contrary, introduces a modern hybrid and much more robust loop detection mechanism. In this method, the initial keypoints are first extracted using the Superpoint model \cite{detone2018superpoint}. 
The keypoints are then filtered using the semantic segmentation neural network of the visual preprocessing component. In this stage, there is no need for checking the fundamental matrix test to add semi-dynamic objects’ keypoints because they may not stay steady in the long-term. 

Subsequent to this stage, the descriptors of keypoints of each keyframe are stored in a DBoW2 \cite{galvez2012bags} database to detect new loops in the future. Whenever a new keyframe comes in, its $K$ closest keyframes from the DBoW2 database are sought using a weak KNN algorithm. If the number of matched descriptors (keypoints) between the input frame and any of these keyframes is more than a predefined threshold ($T_{\textit{KNN}}$), the corresponding pair of keyframes are passed to the Superglue\cite{sarlin2020superglue} matching algorithm. Finally, there is a PnP and pose matching stage to determine the exact transformation between two keyframes. This transformation and the corresponding keyframes' indices are passed to pose graph optimization block to close the detected loop. The Superglue correspondence detection algorithm allows SRVIO to match significant viewpoint changes, occluded scenes, day/night scenarios and many other challenging conditions as demonstrated in \cite{sarlin2020superglue}.

The keyframe database contains the following data:
\begin{align}
\left[i, \hat{\mathbf{p}}_{i}^{w}, \hat{\mathbf{q}}_{i}^{w}, j, \hat{\mathbf{p}}_{i, j}^{i}, \hat{\psi}_{i, j}, \mathbf{D}(u, v, des)\right]
\end{align}

where $\hat{\mathbf{p}}_{i}^{w}$ and $\hat{\mathbf{q}}_{i}^{w}$ are estimated position and orientation of the $i^{th}$ keyframe, respectively. $j$ is the possible loop-closed keyframe index. $\hat{\mathbf{p}}_{i, j}$ and $\hat{\psi}_{i, j}$ are the estimated relative position and yaw angle between these two frames ($i$ and $j$), respectively. $\mathbf{D}(u, v, des)$ is the keypoint matrix of $i^{th}$ keyframe containing pixel locations $(u,v)$ and $des$ descriptors. This data matrix may be considered as the output of the proposed VIO framework.

\begin{figure} 
\centerline{\includegraphics[width=\columnwidth]{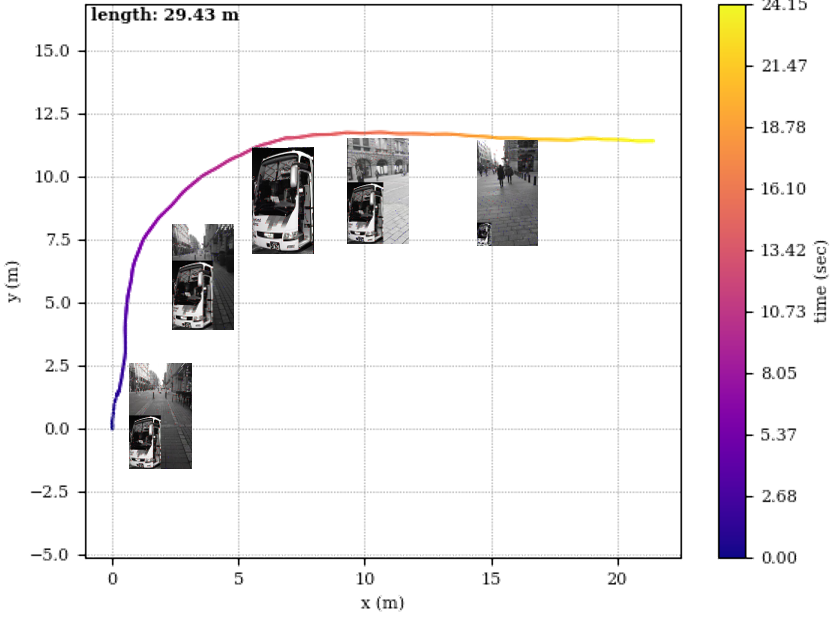}}
\caption{A simulated sequence based on ADVIO Sequence 22 in which a bus gets close to the camera and covers the whole camera view for a few seconds and then goes away. This simulation is designed to mimic the situation when a car is moving behind a bus in traffic.}
\label{fig:advio22}
\end{figure}

\begin{figure} 
\centerline{\includegraphics[width=\columnwidth]{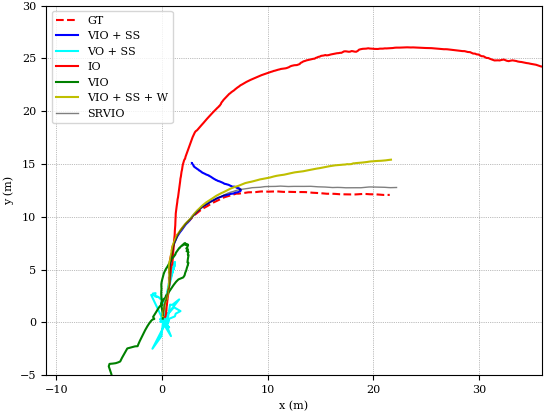}}
\caption{The results of the proposed SRVIO model and its sub-models over the simulated ADVIO sequence 22. The VIO corresponds to a basic VIO framework with no semantic segmentation and denoising networks. In this experiment, the loop-closure mechanism is turned off.}
\label{fig:res_advio22}
\end{figure}

\begin{figure} 
\centerline{\includegraphics[width=\columnwidth]{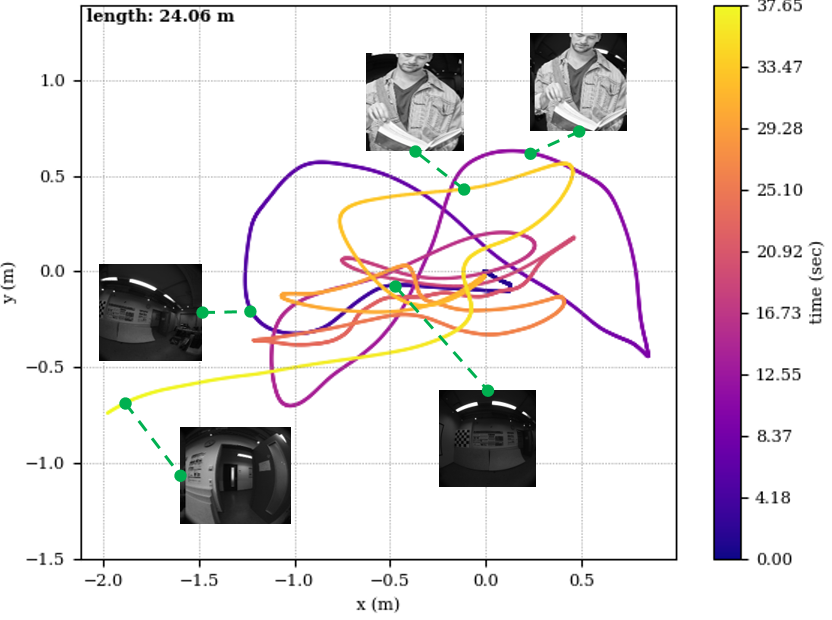}}
\caption{The simulated outdoor7 sequence of the TUM-VI dataset has a human blocking most of the camera's view. The human appears at second 10 and leaves at second 28.}
\label{fig:tumvi7}
\end{figure}

\begin{figure} 
\centerline{\includegraphics[width=\columnwidth]{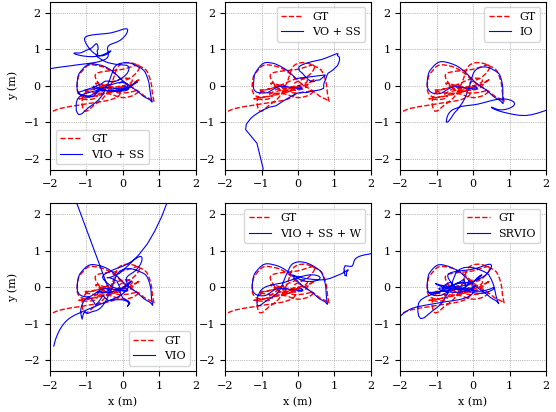}}
\caption{The results of the proposed SRVIO model and its sub-models over the simulated TUM-VI outdoor7 sequence.In this experiment the loop-closure mechanism is turned off.}
\label{fig:res_tumvi7}
\end{figure}

\begin{table}[t] 
\centering
\begin{center}
\begin{small}
\begin{tabular}{|l|c|c|c|c|} 
\hline
Model & Sim1 ATE & Sim1 RPE & Sim2 ATE & Sim2 RPE\\
\hline\hline
VIO                & 22.691 & 0.207  & 5.150 & 2.578\\
VO+SS              & 12.826 & 0.706  & 1.223 & 0.709\\
IO                 & 19.712 & 0.094  & 5.084 & 0.363\\
VIO+SS             & 9.457  & 0.061   & 1.663 & 0.336\\
VIO+SS+W           & 7.548  & 0.026  & 3.709 & 0.348\\
\textbf{SRVIO}     & \textbf{2.201}  & \textbf{0.024}  & \textbf{0.959} & \textbf{0.180}\\
\hline
\end{tabular}
\end{small}
\end{center}
\caption{The results of the proposed SRVIO model and its sub-models over \textbf{ADVIO22-sim (Sim1) and TUM-VI outdoor7-sim (Sim2)}. The error metrics are RMS of ATE and RPE (m). The errors are not divided by path length of 29.8m to have a better distiction with limited floating point numbers.}
\label{table:res_advio22}
\end{table}

\begin{table}[t] 
\centering
\begin{center}
\begin{small}
\begin{tabular}{|l|c|c|c|} 
\hline
Model & s\_half & w\_xyz & w\_rpy\\
\hline\hline
VINS-Mono \cite{qin2018vins}                &  0.043 & 0.478 & 0.601\\
ORB-SLAM3 \cite{campos2021orb}              &  0.019 & 0.202 & 0.515\\
dynamic-SLAM    \cite{xiao2019dynamic}      &  0.017 & 0.465 & 0.540 \\
DS-SLAM  \cite{yu2018ds}                    &  0.016 & 0.021 & 0.330 \\
DynaSLAM \cite{bescos2018dynaslam}          &  0.017 & 0.015 & 0.035\\
SLAMANTIC \cite{schorghuber2019slamantic}   &  0.016 & \textbf{0.016} & 0.043\\
\hline
\textbf{SRVIO}                              & \textbf{0.014} & \textbf{0.016} & \textbf{0.028}\\
\hline
\end{tabular}
\end{small}
\end{center}
\caption{Frameworks evaluated on \textbf{TUM-RGBD dynamic} dataset. Error metric is RMS of ATE / path length.}
\label{table:res_tumdynamic}
\end{table}

\begin{table}[t] 
\centering
\begin{center}
\begin{small}
\begin{tabular}{|l|c|c|c|c|c|} 
\hline
Model & V103 & V202 & V203 & M03 & M05\\
\hline\hline
VINS-Mono   \cite{qin2018vins}   & 0.130 & 0.088 & 0.214 & 0.074 & 0.147\\
VI-DSO      \cite{von2018direct} & 0.096 & 0.062 & 0.174 & 0.117 & 0.121\\
ORB-SLAM3   \cite{campos2021orb}   & 0.025 & \textbf{0.015} & \textbf{0.037} & 0.033 & 0.071\\
UnDeepVO    \cite{li2018undeepvo} & 1.000 & 1.240 & 0.780 & 1.800 & 0.880\\
D3VO        \cite{yang2020d3vo}        & 0.110 & 0.050 & 0.190 & 0.080 & 0.090\\
\hline
\textbf{SRVIO no LC}  & 0.104 & 0.027 & 0.093 & 0.082 & 0.109\\
\textbf{SRVIO}        & \textbf{0.021} & 0.017 & 0.042 & \textbf{0.032} & \textbf{0.062}\\
\hline
\end{tabular}
\end{small}
\end{center}
\caption{Frameworks evaluated on \textbf{EuRoC} dataset. Error metric is RMS of ATE / path length.}
    \label{table:res_EoRoC}
\end{table}

\begin{table}[t] 
\centering
\begin{center}
\begin{small}
\begin{tabular}{|l|c|c|c|c|} 
\hline
Model & Room3 & Room5 & Outdoor6 & Outdoor7\\
\hline\hline
VINS-Mono    & 0.11 & 0.20 & 130.6 & 21.90\\
VI-DSO       & 0.13 & 0.13 & 65.6 & 4.07\\
ORB-SLAM3    & \textbf{0.04} & \textbf{0.02} & 16.84 & 7.59\\
\hline
\textbf{SRVIO}        & 0.06 & 0.08 & \textbf{13.19} & \textbf{3.87}\\
\textbf{SRVIO} (no SS)& 0.05 & 0.06 & 13.73 & 3.90\\
\hline
\end{tabular}
\end{small}
\end{center}
\caption{Frameworks evaluated on \textbf{TUM-VI} dataset. Error metric is RMS of ATE / path length.}
    \label{table:res_tumvi}
\end{table}

\section{Experiments and results}
\label{results}
In this section, multiple experiments are performed to showcase the superiority of SRVIO in the aforementioned challenging conditions over the current state-of-the-art VIO models.

\subsection{Datasets and Parameter setting}
We have selected several datasets to evaluate SRVIO against competitors.
The ADVIO \cite{cortes2018advio} sequence 22 is edited to include highly dynamic objects in order to compare the outcomes of the models in such situations. TUM-RGBD dynamic \cite{sturm12iros} is a standard visual-only dataset. Here, TUM-RGBD is used with some estimators to have IMU data. This modified dataset is then used to evaluate the performance of dynamic SLAM models.
The Simulated TUM-VI \cite{schubert2018tum} outdoor7 sequence is used to evaluate the outcomes of SLAM frameworks in a case in which a dynamic object completely covers the camera view. To do so, blocking objects are added to the corresponding video. Lastly, SRVIO is evaluated on multiple sequences of EuRoC \cite{burri2016euroc} and TUM-VI datasets to demonstrate the SRVIO's performance in normal conditions.

The experiments are performed on a system consisting of an i5-6500 CPU and two GTX-1080ti and GTX-1080 GPUs with a total RAM of 24 GBs and a Samsung 850 SSD drive. At each experiment, if a framework processes less than 10 frames per second, it is assumed that the framework have failed at that experiment. We have also used Ubuntu 16.04 LTS operating system with CUDA toolkit 10.1 and Nvidia driver 450.80, ROS Kinetic, Pytorch 1.4.0, Torchvision 0.5.0 and OpenCV 3.4.2 in our experiments.

The SRVIO's parameters are set as follows. The max feature count and threshold are set to 300 and 150, respectivly. Tracking frequency is chosen to be 20 fps. Max solver time for Ceres solver is set to 12 loops of 0.06 secs. 
As for the neural networks, in order to give ADVIO 1280x720 images to the Superpoint NN for retrieving keypoints, we first resize the images to 512x360 and then add zero padding to make it 512x512. The Superpoint and Superglue are fine tuned on some unused sequences of TUM-VI and EoRoC. The HRNet semantic segmentation model is pretrained on Cityscapes dataset and there are no training on TUM-VI or EuRoC or TUM-RGBD dynamic. The architecture of Superpoint and Superglue and HRNet are not altered. As for the Denoising CNNs, we use 7 layers of dilated CNNs for each of gyro data denoising and accelerometer data denoising. The activation functions are GELU to reduce overfitting. The learning scheduler is Cosine-sceduler. The training optimizer is Adam. To train this NN for TUM-VI and TUM-RGBD datasets, we use the calibration IMU data (almost 111 hrs) provided by dataset owners to carefully calibrate the denoising CNNs over staionary conditions. As for the ADVIO and EoRoC, we use the unused sequences as calibration data to train the network with them.

Lastly, the evaluations are performed with both absolute trajectory error (ATE) and relative pose error (RPE) using our custom EVO evaluator algorithm. 

\subsection{Ablation study}
Goal of this experiment is to demonstrate the task of each block of SRVIO framework in facing extreme dynamic environments. To do so, we study the results of a number of sub-models of SRVIO. In the rest of this section, VIO stands for a vanilla VIO framework (similar to VINS-Mono) with no semantic segmentation, adaptive weighting, IMU denoising networks, or loop-closure. IO, resembling Inertial Odometry, is the same VIO framework with visual residual deleted from the optimization. Similarly, VO is another version of VIO with no inertial residual inside the optimization formula. SS stands for semantic segmentation module, and if used, it means that the framework uses the visual preprocessing of the SRVIO. W stands for dynamic weighting and if used, the W framework uses the optimization block of SRVIO with inertial weights fixed to unity. Sim1 and sim2 refer to simulated ADVIO sequence 22 and TUM-VI sequence outdoor7, respectively.

For the experiments of this section, we have used two sequences whose trajectories are plotted in Fig~\ref{fig:advio22} and Fig.~\ref{fig:tumvi7}. The first sequence (Fig.~\ref{fig:advio22}) is designed to simulate driving behind a bus or walking behind a large object. The second sequence (Fig.~\ref{fig:tumvi7}) is designed to evaluate the robustness of the models against complete coverage of the camera view. In this sequence, he camera view is blocked from second 10 to second 28 by a human. 

The results of SRVIO and its sub-models over the above two sequences are reported in Fig.~\ref{fig:res_advio22}, Fig.~\ref{fig:res_tumvi7} and table \ref{table:res_advio22}. These results indicate that vanilla VIO could never solve the impossible task of odometry on ADVIO22-sim or TUM-VI-sim. 
Furthermore, the IO configuration diverges after a few seconds at both ADVIO22-sim (sim1) and TUM-VI outdoor7-sim (sim2), and therefore the loss is very high compared to other configs (see Table \ref{table:res_advio22}). However, the RPE loss is relatively smaller than expected. This small RPE error is because the IO problem is the build-up error of noise inside inertial integration, and everything seems to be ok in the short term. The VIO configuration also diverges because the semantic segmentation is not used and visual residual becomes large and inertial residual cannot help rescue the framework from divergence. The visual-only model combined with the semantic segmentation network (VO+SS) cannot converge at the sim1 sequence when the bus comes near and covers the image. The same thing happens in the sim2 sequence when the human completely covers the camera. This is because the VO+SS framework does not have enough static keypoints when the bus is near, and the odometry trajectory estimation diverges. The results of the VIO equipped with the semantic segmentation (VIO + SS) are slightly different. This configuration's odometry estimation does not diverge when the bus covers the camera in sim1 sequence. This convergence is because inertial data assists the optimization. However, it diverges after a while because it does not have enough keypoints to track and optimize correctly. In the sim2 sequence, the loss of keypoints and small inertial residual causes the algorithm to maintain movement bias (fixed pose). When the human goes away, the trajectory continues to be accurate.

The VIO combined with semantic segmentation and dynamic weighting inside the optimization block (VIO+SS+W) seems to  perform great at sim1 sequence. Nevertheless, at sim2 sequence, it increase the weight of the inertial residual and hence the divergence of the inertial integration happens. The need for the denoising neural network becomes evident in this experiment. The (VIO+SS+W) configuration shows the importance of having dynamic weighting. It also explains the vulnerability of solely using the weighting mechanism and placing too much attention on inertial data in the optimization step. The complete SRVIO (without loop-closure) is the most successful model in both sim1 and sim2 sequences. It shows how IMU denoising can result in robustness over harsh camera blocking situations and help the VIO framework work more accurately when visual features are not good. 

To summarize, it can be concluded that using only semantic segmentation is not enough and methods similar to SLAMANTIC, DynaSLAM2 would not output accurate trajectories in the extreme dynamic conditions. Also, Other state-of-the-art methods like ORB-SLAM3 and VINS-Mono will not be as accurate as SRVIO in full or partial visual feature loss situations.

\subsection{Comparison with the state-of-the-art models}
In this section, SRVIO's performance is compared with state-of-the-art VIO methods in two main experiments.
In the first experiment, the accuracy of the SRVIO on standard dynamic datasets is evaluated and compared with  state-of-the-art methods. The IMU data of the vision-only dataset are simulated using the ground truth and the additional SLAMANTIC output. The results, reported in Table \ref{table:res_tumdynamic}, indicate that when dynamic objects and some motion blur are present in the input sequences, conventional methods like ORB-SLAM3 and VINS-Mono are useless. Also, the results suggest that our SRVIO method is superior to the SLAMANTIC method even when the IMU data is simulated. 

In the TUM-RGBD dynamic dataset, the walking sequences contain humans walking in front of the camera, and the camera moves rapidly. Also, there are some camera blurs and extreme brightness changes. As the results show (Table \ref{table:res_tumdynamic}), the challenge is relatively high; therefore, the ORB-SLAM3 and VINS-Mono frameworks diverge and give unacceptable results. However, Other dynamic methods use semantic segmentation and thus have much better performance. Although, the other methods have problems in camera blur and extreme light changes, the SRVIO framework compensates the lack of proper visual data using IMU data and the adaptive weighting mechanism. Therefore, the SRVIO is the best choice for indoor datasets with dynamic objects and challenging environments similar to this dataset. Despite having simulated the IMU data, the results at Table \ref{table:res_tumdynamic} show that our method outperforms all previous dynamic methods on this standard dataset.

In the last experiment, SRVIO is compared in normal conditions against state-of-the-art geometric and deep NN SLAM frameworks on EuRoC and TUM-VI datasets. The results are depicted in Table \ref{table:res_EoRoC} and Table \ref{table:res_tumvi}. The results show that pure deep NN SLAM methods like D3VO has comparable results against geometric frameworks with disabled loop closure such as VINS-Mono and VI-DSO. Also, the results show that our SRVIO framework is great under normal conditions as well and its performance does not drop with semantic segmentation as reported by  SLAMANTIC method. The reason behind this is the IMU denoised data which helps to reduce the false semantic segmentation errors on trajectory estimation. Moreover, The ORB-SLAM3 method is excellent on both datasets, and our SRVIO framework has comparable results with this model in the general conditions. Nevertheless, the TUM-VI outdoor sequences show the superiority of our SRVIO framework's loop closure mechanism over the ORB-SLAM3's loop closure block.
To summarize, Our SRVIO framework outperformed all state-of-the-art methods in the dynamic conditions and general conditions. Also, SRVIO in common sequences with easy or no loop closure conditions had comparable results with the state-of-the-art ORB-SLAM3 framework.



\section{Conclusions}
\label{conclusion}
As mentioned, there are three main unsolved challenges in existing state-of-the-art VIO/VO SLAM frameworks: dynamic objects especially camera blocking objects, the divergence of IMU data odometry, and robust long-term loop-closure in challenging conditions such as day/night and multi-view situations. Our proposed SRVIO framework is demonstrated to solve all of the problems at once and become the best choice in those challenging conditions. Moreover, our proposed framework is able to run on any dataset with framerates higher than 10fps with the mentioned computational power.

Our experiments on simulated data also show the need for a dynamic and challenging dataset. There is a simulated dataset for this purpose called VIODE\cite{minoda2021viode}, but this dataset is not real and may cause some biases to hybrid or deep-NN-based SLAM frameworks. There should be a dataset with temporally missing visual and inertial data, dynamic objects,  challenging loop-closure scenes, and scenarios in which the visual features are not trackable (e.g. extreme motion blur). Such dataset may facilitate the design of super robust SLAM frameworks.



%
\bibliographystyle{IEEEtran}
\bibliography{IEEEabrv,main}

\begin{thebibliography}{10}
\providecommand{\url}[1]{#1}
\csname url@samestyle\endcsname
\providecommand{\newblock}{\relax}
\providecommand{\bibinfo}[2]{#2}
\providecommand{\BIBentrySTDinterwordspacing}{\spaceskip=0pt\relax}
\providecommand{\BIBentryALTinterwordstretchfactor}{4}
\providecommand{\BIBentryALTinterwordspacing}{\spaceskip=\fontdimen2\font plus
\BIBentryALTinterwordstretchfactor\fontdimen3\font minus
  \fontdimen4\font\relax}
\providecommand{\BIBforeignlanguage}[2]{{%
\expandafter\ifx\csname l@#1\endcsname\relax
\typeout{** WARNING: IEEEtran.bst: No hyphenation pattern has been}%
\typeout{** loaded for the language `#1'. Using the pattern for}%
\typeout{** the default language instead.}%
\else
\language=\csname l@#1\endcsname
\fi
#2}}
\providecommand{\BIBdecl}{\relax}
\BIBdecl

\bibitem{an2017semantic}
L.~An, X.~Zhang, H.~Gao, and Y.~Liu, ``Semantic segmentation--aided visual
  odometry for urban autonomous driving,'' \emph{International Journal of
  Advanced Robotic Systems}, vol.~14, no.~5, p. 1729881417735667, 2017.

\bibitem{sabry2019ground}
M.~Sabry, A.~Al-Kaff, A.~Hussein, and S.~Abdennadher, ``Ground vehicle
  monocular visual odometry,'' in \emph{2019 IEEE Intelligent Transportation
  Systems Conference (ITSC)}.\hskip 1em plus 0.5em minus 0.4em\relax IEEE,
  2019, pp. 3587--3592.

\bibitem{lin2018autonomous}
Y.~Lin, F.~Gao, T.~Qin, W.~Gao, T.~Liu, W.~Wu, Z.~Yang, and S.~Shen,
  ``Autonomous aerial navigation using monocular visual-inertial fusion,''
  \emph{Journal of Field Robotics}, vol.~35, no.~1, pp. 23--51, 2018.

\bibitem{delmerico2018benchmark}
J.~Delmerico and D.~Scaramuzza, ``A benchmark comparison of monocular
  visual-inertial odometry algorithms for flying robots,'' in \emph{2018 IEEE
  International Conference on Robotics and Automation (ICRA)}.\hskip 1em plus
  0.5em minus 0.4em\relax IEEE, 2018, pp. 2502--2509.

\bibitem{do2019high}
T.~Do, L.~C. Carrillo-Arce, and S.~I. Roumeliotis, ``High-speed autonomous
  quadrotor navigation through visual and inertial paths,'' \emph{The
  International Journal of Robotics Research}, vol.~38, no.~4, pp. 486--504,
  2019.

\bibitem{wu2015square}
K.~Wu, A.~Ahmed, G.~A. Georgiou, and S.~I. Roumeliotis, ``A square root inverse
  filter for efficient vision-aided inertial navigation on mobile devices.'' in
  \emph{Robotics: Science and Systems}, vol.~2, 2015.

\bibitem{qin2018vins}
T.~Qin, P.~Li, and S.~Shen, ``Vins-mono: A robust and versatile monocular
  visual-inertial state estimator,'' \emph{IEEE Transactions on Robotics},
  vol.~34, no.~4, pp. 1004--1020, 2018.

\bibitem{von2018direct}
L.~Von~Stumberg, V.~Usenko, and D.~Cremers, ``Direct sparse visual-inertial
  odometry using dynamic marginalization,'' in \emph{2018 IEEE International
  Conference on Robotics and Automation (ICRA)}.\hskip 1em plus 0.5em minus
  0.4em\relax IEEE, 2018, pp. 2510--2517.

\bibitem{xiong2019ds}
X.~Xiong, W.~Chen, Z.~Liu, and Q.~Shen, ``Ds-vio: Robust and efficient stereo
  visual inertial odometry based on dual stage ekf,'' \emph{arXiv preprint
  arXiv:1905.00684}, 2019.

\bibitem{campos2021orb}
C.~Campos, R.~Elvira, J.~J.~G. Rodr{\'\i}guez, J.~M. Montiel, and J.~D.
  Tard{\'o}s, ``Orb-slam3: An accurate open-source library for visual,
  visual--inertial, and multimap slam,'' \emph{IEEE Transactions on Robotics},
  2021.

\bibitem{clark2017vinet}
R.~Clark, S.~Wang, H.~Wen, A.~Markham, and N.~Trigoni, ``Vinet: Visual-inertial
  odometry as a sequence-to-sequence learning problem,'' in \emph{Proceedings
  of the AAAI Conference on Artificial Intelligence}, vol.~31, no.~1, 2017.

\bibitem{yang2020d3vo}
N.~Yang, L.~v. Stumberg, R.~Wang, and D.~Cremers, ``D3vo: Deep depth, deep pose
  and deep uncertainty for monocular visual odometry,'' in \emph{Proceedings of
  the IEEE/CVF Conference on Computer Vision and Pattern Recognition}, 2020,
  pp. 1281--1292.

\bibitem{nister2004efficient}
D.~Nist{\'e}r, ``An efficient solution to the five-point relative pose
  problem,'' \emph{IEEE transactions on pattern analysis and machine
  intelligence}, vol.~26, no.~6, pp. 756--770, 2004.

\bibitem{schonberger2016structure}
J.~L. Schonberger and J.-M. Frahm, ``Structure-from-motion revisited,'' in
  \emph{Proceedings of the IEEE conference on computer vision and pattern
  recognition}, 2016, pp. 4104--4113.

\bibitem{mur2017orb}
R.~Mur-Artal and J.~D. Tard{\'o}s, ``Orb-slam2: An open-source slam system for
  monocular, stereo, and rgb-d cameras,'' \emph{IEEE Transactions on Robotics},
  vol.~33, no.~5, pp. 1255--1262, 2017.

\bibitem{bescos2018dynaslam}
B.~Bescos, J.~M. F{\'a}cil, J.~Civera, and J.~Neira, ``Dynaslam: Tracking,
  mapping, and inpainting in dynamic scenes,'' \emph{IEEE Robotics and
  Automation Letters}, vol.~3, no.~4, pp. 4076--4083, 2018.

\bibitem{shi1994good}
J.~Shi \emph{et~al.}, ``Good features to track,'' in \emph{1994 Proceedings of
  IEEE conference on computer vision and pattern recognition}.\hskip 1em plus
  0.5em minus 0.4em\relax IEEE, 1994, pp. 593--600.

\bibitem{bouguet2001pyramidal}
J.-Y. Bouguet \emph{et~al.}, ``Pyramidal implementation of the affine lucas
  kanade feature tracker description of the algorithm,'' \emph{Intel
  corporation}, vol.~5, no. 1-10, p.~4, 2001.

\bibitem{li2018undeepvo}
R.~Li, S.~Wang, Z.~Long, and D.~Gu, ``Undeepvo: Monocular visual odometry
  through unsupervised deep learning,'' in \emph{2018 IEEE international
  conference on robotics and automation (ICRA)}.\hskip 1em plus 0.5em minus
  0.4em\relax IEEE, 2018, pp. 7286--7291.

\bibitem{liu2019neural}
C.~Liu, J.~Gu, K.~Kim, S.~G. Narasimhan, and J.~Kautz, ``Neural rgb (r) d
  sensing: Depth and uncertainty from a video camera,'' in \emph{Proceedings of
  the IEEE/CVF Conference on Computer Vision and Pattern Recognition}, 2019,
  pp. 10\,986--10\,995.

\bibitem{mur2015orb}
R.~Mur-Artal, J.~M.~M. Montiel, and J.~D. Tardos, ``Orb-slam: a versatile and
  accurate monocular slam system,'' \emph{IEEE transactions on robotics},
  vol.~31, no.~5, pp. 1147--1163, 2015.

\bibitem{klein2007parallel}
G.~Klein and D.~Murray, ``Parallel tracking and mapping for small ar
  workspaces,'' in \emph{2007 6th IEEE and ACM international symposium on mixed
  and augmented reality}.\hskip 1em plus 0.5em minus 0.4em\relax IEEE, 2007,
  pp. 225--234.

\bibitem{kaneko2018mask}
M.~Kaneko, K.~Iwami, T.~Ogawa, T.~Yamasaki, and K.~Aizawa, ``Mask-slam: Robust
  feature-based monocular slam by masking using semantic segmentation,'' in
  \emph{Proceedings of the IEEE Conference on Computer Vision and Pattern
  Recognition Workshops}, 2018, pp. 258--266.

\bibitem{yu2018ds}
C.~Yu, Z.~Liu, X.-J. Liu, F.~Xie, Y.~Yang, Q.~Wei, and Q.~Fei, ``Ds-slam: A
  semantic visual slam towards dynamic environments,'' in \emph{2018 IEEE/RSJ
  International Conference on Intelligent Robots and Systems (IROS)}.\hskip 1em
  plus 0.5em minus 0.4em\relax IEEE, 2018, pp. 1168--1174.

\bibitem{han2020dynamic}
S.~Han and Z.~Xi, ``Dynamic scene semantics slam based on semantic
  segmentation,'' \emph{IEEE Access}, vol.~8, pp. 43\,563--43\,570, 2020.

\bibitem{zhao2017pyramid}
H.~Zhao, J.~Shi, X.~Qi, X.~Wang, and J.~Jia, ``Pyramid scene parsing network,''
  in \emph{Proceedings of the IEEE conference on computer vision and pattern
  recognition}, 2017, pp. 2881--2890.

\bibitem{li2017rgb}
S.~Li and D.~Lee, ``Rgb-d slam in dynamic environments using static point
  weighting,'' \emph{IEEE Robotics and Automation Letters}, vol.~2, no.~4, pp.
  2263--2270, 2017.

\bibitem{xiao2019dynamic}
L.~Xiao, J.~Wang, X.~Qiu, Z.~Rong, and X.~Zou, ``Dynamic-slam: Semantic
  monocular visual localization and mapping based on deep learning in dynamic
  environment,'' \emph{Robotics and Autonomous Systems}, vol. 117, pp. 1--16,
  2019.

\bibitem{woodman2007introduction}
O.~J. Woodman, ``An introduction to inertial navigation,'' University of
  Cambridge, Computer Laboratory, Tech. Rep., 2007.

\bibitem{savage1998strapdown}
P.~G. Savage, ``Strapdown inertial navigation integration algorithm design part
  1: Attitude algorithms,'' \emph{Journal of guidance, control, and dynamics},
  vol.~21, no.~1, pp. 19--28, 1998.

\bibitem{chen2018ionet}
C.~Chen, X.~Lu, A.~Markham, and N.~Trigoni, ``Ionet: Learning to cure the curse
  of drift in inertial odometry,'' in \emph{Proceedings of the AAAI Conference
  on Artificial Intelligence}, vol.~32, no.~1, 2018.

\bibitem{brossard2020denoising}
M.~Brossard, S.~Bonnabel, and A.~Barrau, ``Denoising imu gyroscopes with deep
  learning for open-loop attitude estimation,'' \emph{IEEE Robotics and
  Automation Letters}, vol.~5, no.~3, pp. 4796--4803, 2020.

\bibitem{gawel2018x}
A.~Gawel, C.~Del~Don, R.~Siegwart, J.~Nieto, and C.~Cadena, ``X-view:
  Graph-based semantic multi-view localization,'' \emph{IEEE Robotics and
  Automation Letters}, vol.~3, no.~3, pp. 1687--1694, 2018.

\bibitem{sarlin2020superglue}
P.-E. Sarlin, D.~DeTone, T.~Malisiewicz, and A.~Rabinovich, ``Superglue:
  Learning feature matching with graph neural networks,'' in \emph{Proceedings
  of the IEEE/CVF conference on computer vision and pattern recognition}, 2020,
  pp. 4938--4947.

\bibitem{yuan2020object}
Y.~Yuan, X.~Chen, and J.~Wang, ``Object-contextual representations for semantic
  segmentation,'' in \emph{Computer Vision--ECCV 2020: 16th European
  Conference, Glasgow, UK, August 23--28, 2020, Proceedings, Part VI 16}.\hskip
  1em plus 0.5em minus 0.4em\relax Springer, 2020, pp. 173--190.

\bibitem{schorghuber2019slamantic}
M.~Schorghuber, D.~Steininger, Y.~Cabon, M.~Humenberger, and M.~Gelautz,
  ``Slamantic-leveraging semantics to improve vslam in dynamic environments,''
  in \emph{Proceedings of the IEEE International Conference on Computer Vision
  Workshops}, 2019, pp. 0--0.

\bibitem{fischler1981random}
M.~A. Fischler and R.~C. Bolles, ``Random sample consensus: a paradigm for
  model fitting with applications to image analysis and automated
  cartography,'' \emph{Communications of the ACM}, vol.~24, no.~6, pp.
  381--395, 1981.

\bibitem{detone2018superpoint}
D.~DeTone, T.~Malisiewicz, and A.~Rabinovich, ``Superpoint: Self-supervised
  interest point detection and description,'' in \emph{Proceedings of the IEEE
  conference on computer vision and pattern recognition workshops}, 2018, pp.
  224--236.

\bibitem{galvez2012bags}
D.~G{\'a}lvez-L{\'o}pez and J.~D. Tardos, ``Bags of binary words for fast place
  recognition in image sequences,'' \emph{IEEE Transactions on Robotics},
  vol.~28, no.~5, pp. 1188--1197, 2012.

\bibitem{cortes2018advio}
S.~Cort{\'e}s, A.~Solin, E.~Rahtu, and J.~Kannala, ``Advio: An authentic
  dataset for visual-inertial odometry,'' in \emph{Proceedings of the European
  Conference on Computer Vision (ECCV)}, 2018, pp. 419--434.

\bibitem{sturm12iros}
J.~Sturm, N.~Engelhard, F.~Endres, W.~Burgard, and D.~Cremers, ``A benchmark
  for the evaluation of rgb-d slam systems,'' in \emph{Proc. of the
  International Conference on Intelligent Robot Systems (IROS)}, Oct. 2012.

\bibitem{schubert2018tum}
D.~Schubert, T.~Goll, N.~Demmel, V.~Usenko, J.~St{\"u}ckler, and D.~Cremers,
  ``The tum vi benchmark for evaluating visual-inertial odometry,'' in
  \emph{2018 IEEE/RSJ International Conference on Intelligent Robots and
  Systems (IROS)}.\hskip 1em plus 0.5em minus 0.4em\relax IEEE, 2018, pp.
  1680--1687.

\bibitem{burri2016euroc}
M.~Burri, J.~Nikolic, P.~Gohl, T.~Schneider, J.~Rehder, S.~Omari, M.~W.
  Achtelik, and R.~Siegwart, ``The euroc micro aerial vehicle datasets,''
  \emph{The International Journal of Robotics Research}, vol.~35, no.~10, pp.
  1157--1163, 2016.

\bibitem{minoda2021viode}
K.~Minoda, F.~Schilling, V.~W{\"u}est, D.~Floreano, and T.~Yairi, ``Viode: A
  simulated dataset to address the challenges of visual-inertial odometry in
  dynamic environments,'' \emph{IEEE Robotics and Automation Letters}, vol.~6,
  no.~2, pp. 1343--1350, 2021.

\end{thebibliography}

\end{document}